\documentclass[10pt,twocolumn,letterpaper]{article}

\usepackage{cvpr} 
\usepackage{times}
\usepackage{epsfig}
\usepackage{graphicx}
\usepackage{amsmath,amssymb}
\usepackage[numbers,sort&compress]{natbib}
\usepackage{tabularray}
\usepackage{adjustbox}
\usepackage{doi}
\usepackage{hyperref}

\usepackage{caption}
\begin{document}

\title{Towards Robust Iris Recognition Through Occlusion Identification and Conditional Diffusion-Based Reconstruction}

\author{
Kamrul Hasan, Mylene C.Q. Farias, Oleg V. Komogortsev\\
Texas State University, San Marcos, Texas, USA\\
{\tt\small \{kamrul.hasan, mylene, ok\}@txstate.edu}
}


\maketitle
\thispagestyle{empty}

\begin{abstract}
Iris recognition is a reliable biometric approach that identifies individuals using the distinctive and stable texture of the iris. However, recognition performance can degrade when discriminative iris texture is partially occluded by eyelids, eyelashes, specular reflections, or other acquisition artifacts. Existing approaches often perform recognition directly on degraded samples or rely only on the remaining visible iris region, which may be inadequate when substantial texture is corrupted. To address this limitation, we propose an occlusion-aware iris recognition framework with three sequential modules: occlusion-type identification, diffusion-based reconstruction, and deep-learning-based recognition. First, a residual 2D CNN-based network determines whether an iris image is non-occluded or belongs to one of the controlled occlusion categories. Second, the occluded image, binary mask, and predicted occlusion type condition a denoising diffusion probabilistic model to reconstruct the corrupted region. Finally, VGG19-HPMNet, a modified VGG19 model with horizontal pyramid mapping, extracts discriminative global and part-wise local iris features for recognition. Experiments on the CASIA-Iris-Thousand dataset under a controlled synthetic-occlusion protocol show that the proposed framework improves iris recognition performance by identifying the occlusion type, reconstructing masked regions, and re-evaluating the restored iris samples.

\end{abstract}

\section{Introduction}
Biometrics refers to the measurement and analysis of distinctive physiological or behavioral features for individual recognition \cite{jain2011introduction}. Among different biometric traits, such as face \cite{zhao2003face}, fingerprint \cite{maltoni2009handbook}, voice \cite{markowitz2000voice}, and gait \cite{uddin2024horizontal}, iris \cite{daugman2007new} is regarded as one of the most reliable modalities because of its rich texture, distinctiveness, and long-term stability. As a result, iris recognition has been widely adopted as a highly secure authentication approach \cite{daugman2009iris, wildes2002iris}. More recently, deep learning has further advanced iris-based biometrics by improving several core tasks, including localization, segmentation, recognition, presentation attack detection, and cross-condition matching \cite{sundaram2011fast, arsalan2017deep, nguyen2024deep}.

Despite this progress, robust iris recognition remains challenging in practical environments \cite{thavalengal2015iris}. Many iris recognition systems still perform well under controlled acquisition settings that require user cooperation and favorable imaging conditions \cite{harakannanavar2019extensive}. In real-world scenarios, iris images are often occluded and degraded by eyelids, eyelashes, shadows, motion blur, specular reflections, sensor noise, and off-angle capture \cite{hajari2015review}. These factors are more apparent in unconstrained acquisition settings, where subject movement, distance variation, illumination changes, and other environmental conditions directly affect visible iris texture and, consequently, lower recognition performance \cite{benalcazar2019iris, malgheet2021iris, fancourt2005iris}.

Therefore, a common strategy for addressing occlusion is to ignore corrupted regions or extract features only from the remaining visible iris area \cite{li2012automatic}. Although this may work for mildly degraded samples, it becomes unreliable when substantial discriminative texture is missing. To address this issue, several studies \cite{lee2019conditional, mostofa2021deep, khan2023deformirisnet, deng2025high} have explored iris restoration and inpainting. For example, generative adversarial network (GAN)-based methods, including DCGAN, TT-GAN \cite{chen2024two}, and TSDRA-GAN \cite{chen2023coarse}, reconstruct incomplete iris images or textures obscured by eyelids and eyelashes, while I3FDM \cite{li2024i3fdm} recently introduced a diffusion-based iris inpainting framework. However, these methods show promising reconstruction capabilities; they generally focus on restoring degraded images without explicitly identifying the type of occlusion or performing selective reconstruction, which limits their practicality.

Motivated by this research gap, we propose an occlusion-aware iris recognition framework comprising three modules: an occlusion identifier, a diffusion-based reconstructor, and a deep recognition network. The first module consists of residual 2D CNNs that determine whether an input image belongs to a non-occluded class or to one of the controlled-occlusion categories. The second module reconstructs the corrupted region using a denoising diffusion probabilistic model (DDPM) \cite{ho2020denoising} conditioned on the occluded image, the binary mask, the diffusion timestep, and the predicted occlusion class. The third module performs individual recognition using a modified VGG-19 \cite{simonyan2014very} backbone augmented with horizontal pyramid mapping (VGG19-HPMNet), enabling the extraction of both global and part-wise discriminative iris representations. Through this design, the proposed framework evaluates whether explicit occlusion detection and occlusion-guided reconstruction can improve downstream recognition under controlled synthetic occlusion.
The contributions of this paper are summarized as follows:

\begin{enumerate}
\item We propose an occlusion-aware iris recognition framework that integrates occlusion-type identification, conditional diffusion-based reconstruction, and iris recognition within a sequential pipeline.

\item We design a residual 2D CNN-based occlusion identifier that distinguishes non-occluded iris samples from multiple occlusion categories and provides class-conditioning information to the reconstruction module.

\item We introduce an occlusion-guided selective reconstruction module that processes only occluded samples using a conditional DDPM, enabling the restoration process to adapt to the detected occlusion type and improving reconstruction quality.

\item We incorporate horizontal pyramid mapping (HPM) with a VGG-19 model and demonstrate improved downstream biometric performance through reconstruction, recognition, and ablation analyses on the CASIA-Iris-Thousand \cite{casia2017} dataset.
\end{enumerate}

\section{Related Work}
\subsection{Iris Recognition under Occluded Conditions}
Although iris recognition is highly accurate under controlled acquisition, its performance degrades in non-ideal environments due to eyelid/eyelash occlusion, eyeglass obstruction, specular reflections, off-angle capture, blur, and partial iris visibility. Prior studies \cite{poursaberi2006iris, tan2010efficient, li2021robust, alrifaee2024adaptive, wang2020towards} have therefore emphasized robust segmentation and recognition under non-cooperative conditions, showing that accurate preprocessing is critical for unconstrained acquisition. For example, Tan et al. \cite{tan2010efficient} improved recognition using clustering-based localization, an enhanced integrodifferential operator, and learned eyelid/eyelash models, while Wang et al. \cite{wang2020towards} jointly predicted iris masks and boundaries using multitask deep learning with an attention module. Although these approaches improve robustness, they primarily suppress, exclude, or down-weight corrupted pixels rather than reconstructing missing iris texture. This motivates frameworks that explicitly model occlusion and evaluate whether reconstruction can restore useful biometric information.

\subsection{Iris Restoration and Inpainting for Biometric Utility}
Recently, several studies \cite{lee2019conditional, mostofa2021deep, lee2021enhanced, tanna2025occlusionnetplusplus} have explored iris restoration and inpainting to recover missing texture before biometric matching. GAN-based methods have been widely used for this purpose: Zeng et al. \cite{zeng2021incomplete} restored obscured iris regions to improve recognition, TT-GAN \cite{chen2024two} used a two-stage, two-discriminator design for incomplete iris images, and TSDRA-GAN \cite{chen2023coarse} adopted a coarse-to-fine residual attention strategy for textures obscured by eyelids and eyelashes. More recent methods have introduced stronger priors and diffusion models: Gformer \cite{huang2023generative} combines transformer-based encoding with a generative iris prior, while I3FDM \cite{li2024i3fdm} iteratively reconstructs occluded iris images using an inverse fusion module. Although these methods improve iris inpainting quality, they primarily focus on complete image restoration and do not explicitly integrate occlusion identification, selective reconstruction, and downstream biometric recognition.

\section{Proposed Framework}
\subsection{Overview}

Fig.~\ref{fig:main_architecture} shows the proposed three-stage framework. Given an input iris image, the first stage determines whether it is non-occluded or belongs to a specific occlusion class. The second stage uses occlusion information and the occluded image to reconstruct only the corrupted region with a conditional DDPM. The final stage uses VGG19-HPMNet to extract global and part-wise local features for recognition.

\subsection{Data Preprocessing}
CASIA-Iris-Thousand \cite{casia2017} images are relatively large and include a broad periocular region, whereas iris recognition requires only the iris texture. Therefore, each image is loaded at its original grayscale resolution, and then a resized copy is used only for U-Net-based iris segmentation. The predicted mask is mapped back to the original image size, refined with morphological operations, and used to define a margin-expanded bounding box for cropping the centered iris region from the original image. 
The cropped iris Region of Interest (ROI) is then resized to $224 \times 224$ and used as input to the proposed framework.

\subsection{Occlusion Identification Network}
\begin{figure*}[ht]
    \centering
\includegraphics[width=0.8\linewidth]{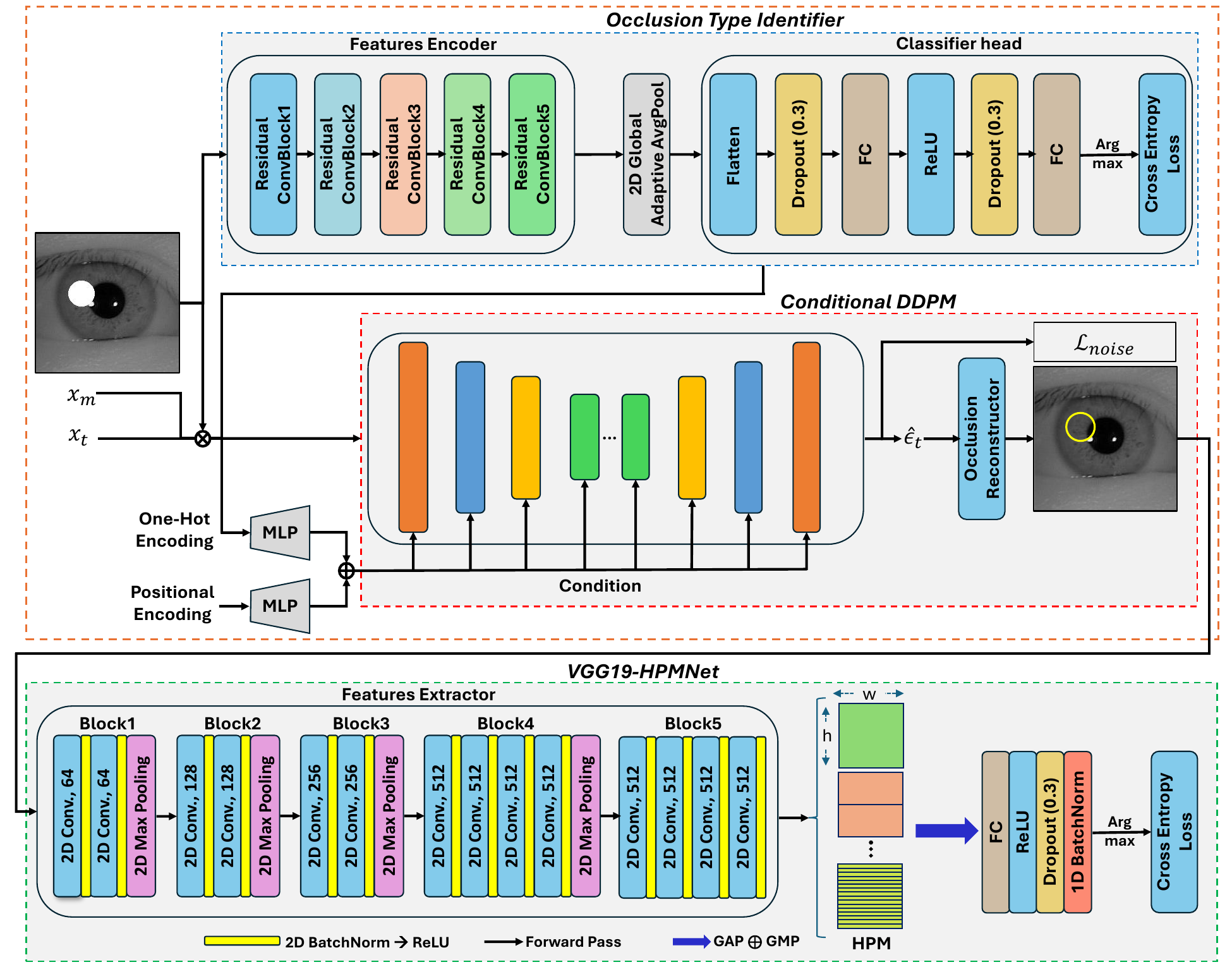}
    \caption{Overview of the proposed occlusion-aware framework, consisting of an occlusion type identifier, a diffusion-based occlusion reconstructor, and VGG19-HPMNet for iris recognition. FC denotes a fully connected layer. In the conditional DDPM module, $x_t$ is the noisy iris image at timestep $t$, $x_m$ is the binary occlusion mask, and $\hat{\varepsilon}_t$ is the predicted noise. MLP, $\otimes$, and $\oplus$ denote a multilayer perceptron,  channel-wise concatenation, and element-wise addition, respectively. In the VGG19-HPMNet module, GAP, GMP, and HPM denote global average pooling, global max pooling, and horizontal pyramid mapping, respectively.}
    \label{fig:main_architecture}
\end{figure*}

The occlusion identification module determines whether an input iris image is non-occluded or belongs to a specific occlusion type. As shown in Fig.~\ref{fig:main_architecture}, it consists of a feature encoder, global average pooling, and a classifier head.

The feature encoder contains five stacked \texttt{ResidualConvBlock} units that increase the channel depth from $1{\to}32{\to}64{\to}128{\to}256{\to}512$ while halving the spatial resolution at each stage. Each block applies two $3{\times}3$ convolutions, adds a residual skip connection using a $1{\times}1$ projection when needed, and then applies $2{\times}2$ pooling, batch normalization (BN), and ReLU activation. The residual addition is performed before pooling so that the skip path operates at full spatial resolution.

The resulting feature map is then reduced to a 512-dimensional vector using 2D global adaptive average pooling. Later, the classifier head applies flattening, dropout, and fully connected layers to produce occlusion-class logits, where $c_0$ denotes no occlusion and $c_1,\ldots,c_{n}$ denote distinct occlusion types. The selected occlusion label is encoded as a one-hot vector and used as conditioning input for the downstream diffusion module. This network is optimized with cross-entropy loss.


\subsection{Conditional Diffusion-Based Reconstruction}
To reconstruct occluded iris images, we employ a conditional denoising diffusion probabilistic model (DDPM) \cite{ho2020denoising} in which the clean image \(x_c\) serves as the diffusion target and the occluded image \(x_o\) provides spatial guidance. Following the standard DDPM forward process, a noisy sample at diffusion step \(t\) is generated as follows:
\begin{equation}
x_t = \sqrt{\bar{\alpha}_t}\,x_c + \sqrt{1-\bar{\alpha}_t}\,\epsilon,
\qquad \epsilon \sim \mathcal{N}(0,I),
\end{equation}
where \(x_t\) denotes the noisy version of the clean image at step \(t\), \(\bar{\alpha}_t=\prod_{s=1}^{t}\alpha_s\) is the cumulative noise-retention factor, and \(\epsilon\) is Gaussian noise. In our implementation, \(x_c\) is first normalized to \([-1,1]\), after which a random timestep \(t\) and noise realization \(\epsilon\) are sampled during training.

Later, a binary occlusion mask \(x_m\) is derived directly from the paired occluded and clean images:
\begin{equation}
x_m(i,j)=
\begin{cases}
1, & \text{if } |x_o(i,j)-x_c(i,j)|>\tau,\\
0, & \text{otherwise},
\end{cases}
\end{equation}
where the absolute difference is computed on the original 8-bit pixel-intensity scale with a fixed threshold $\tau=10$; $x_m=1$ and $x_m=0$ denote corrupted and preserved pixels, respectively. Since $x_m$ is derived from the paired clean reference and synthetically occluded image, the reported experiments use reference-based mask localization.

In addition to spatial guidance, diffusion receives occlusion types as a condition where $c_0,\ldots,c_{n}$ denote the occlusion class label, which is encoded as a one-hot vector as follows:
\begin{equation}
d_c=\mathrm{onehot}(c)\in\{0,1\}^{n},
\end{equation}
where $c$ is the predicted occlusion-class index and $n$ is the number of occlusion classes.

Simultaneously, the diffusion timestep \(t\) is mapped to a sinusoidal positional embedding,
\begin{equation}
d_t = \mathrm{PE}(t),
\end{equation}
where \(\mathrm{PE}(\cdot)\) is the standard sine-cosine timestep encoding followed by a small multilayer perceptron (MLP). The one-hot class vector \(d_c\) is also projected into the same embedding space through a learnable MLP. The final conditioning embedding is then obtained by element-wise summation:
\begin{equation}
d = f_t(d_t) + f_c(d_c),
\end{equation}
where \(f_t(\cdot)\) and \(f_c(\cdot)\) denote the learnable projections for timestep and class conditioning, respectively.

The diffusion U-Net receives both spatial and global conditions. Specifically, the noisy image, occluded image, and binary mask are concatenated channel-wise, which forms a three-channel input to the denoiser as follows:
\begin{equation}
U_t = \mathrm{concat}(x_t, x_o, x_m),
\end{equation}

In addition, the merged embedding \(d\) is injected into each residual block through a learned linear projection and broadcast addition. Therefore, noise prediction is performed as follows:

\begin{equation}
\hat{\epsilon}_t = \epsilon_{\theta}(U_t,t,d)
= \epsilon_{\theta}\!\big(\mathrm{concat}(x_t,x_o,x_m),\, t,\, d\big),
\end{equation}
where \(\epsilon_{\theta}\) denotes the conditional U-Net parameterized by \(\theta\).

During reverse diffusion, the model estimates the mean of the reverse transition using the predicted noise:
\begin{equation}
\mu_{\theta}(x_t,t,d)
=
\frac{1}{\sqrt{\alpha_t}}
\left(
x_t-\frac{1-\alpha_t}{\sqrt{1-\bar{\alpha}_t}}\hat{\epsilon}_t
\right),
\end{equation}
and samples the previous state as follows:
\begin{equation}
x_{t-1}=\mu_{\theta}(x_t,t,d)+\sigma_t z,\qquad z\sim\mathcal{N}(0,I),
\end{equation}
with variance determined by the diffusion schedule. In our implementation, after each reverse step the generated image is composited with the known non-occluded region so that only masked areas are restored:
\begin{equation}
x_{t-1}^{\ast}=x_m\odot x_{t-1} + (1-x_m)\odot \tilde{x}_{o,t-1},
\end{equation}
where \(\tilde{x}_{o,t-1}\) denotes the noised version of the occluded image at step \(t-1\). This repaint-style composition preserves unaffected iris regions while allowing the model to synthesize only the occluded content. Overall, the model is trained using the standard DDPM noise-prediction objective:
\begin{equation}
\mathcal{L}_{\mathrm{noise}}=\left\|\epsilon-\hat{\epsilon}_t\right\|_2^2.
\end{equation}


\subsection{VGG19-HPMNet}

In the recognition stage, VGG19-HPMNet combines VGG-19 \cite{simonyan2014very} with horizontal pyramid mapping (HPM) \cite{fu2019horizontal}. VGG-19 is suitable for iris recognition because its stacked $3\times3$ convolutions capture fine-grained texture patterns. We retain only the convolutional feature extractor and remove the final max-pooling and classifier layers; therefore, a $224\times224$ input produces a $512\times14\times14$ feature map, preserving local spatial information before part-based aggregation. HPM horizontally partitions this feature map using pyramid bins $\{1,2,7,14\}$, where the 1-bin level captures global iris information and the 2-, 7-, and 14-bin levels capture part-wise features. Although HPM was originally introduced for person re-identification, its part-based representation is useful for iris biometrics, as identity information may remain in the visible horizontal bands when other regions are occluded or reconstructed imperfectly.
Within each stripe, horizontal pyramid pooling combines global average pooling (GAP) and global max pooling (GMP): GAP captures the overall stripe response, while GMP emphasizes salient activations. Concatenating pooled features from all pyramid levels forms a global-local descriptor, which is passed through a fully connected layer to obtain a 512-dimensional embedding, followed by additional layers and an identity classifier. The network is trained with softmax cross-entropy loss. During inference, the embedding is $\ell_2$-normalized and used for verification and identification.

\section{Experiments}
\subsection{Dataset}
We conduct our experiments on the publicly available CASIA-Iris-Thousand dataset \cite{casia2017}. It contains 20,000 iris images collected from 1,000 subjects using the IKEMB-100 dual-eye iris camera developed by {IrisKing}\footnote{\url{http://www.irisking.com/}}. The images are acquired under near-infrared imaging conditions, and the dataset is widely used for large-scale iris recognition research. 

\begin{figure}[ht]
    \centering
\includegraphics[width=0.6\linewidth]{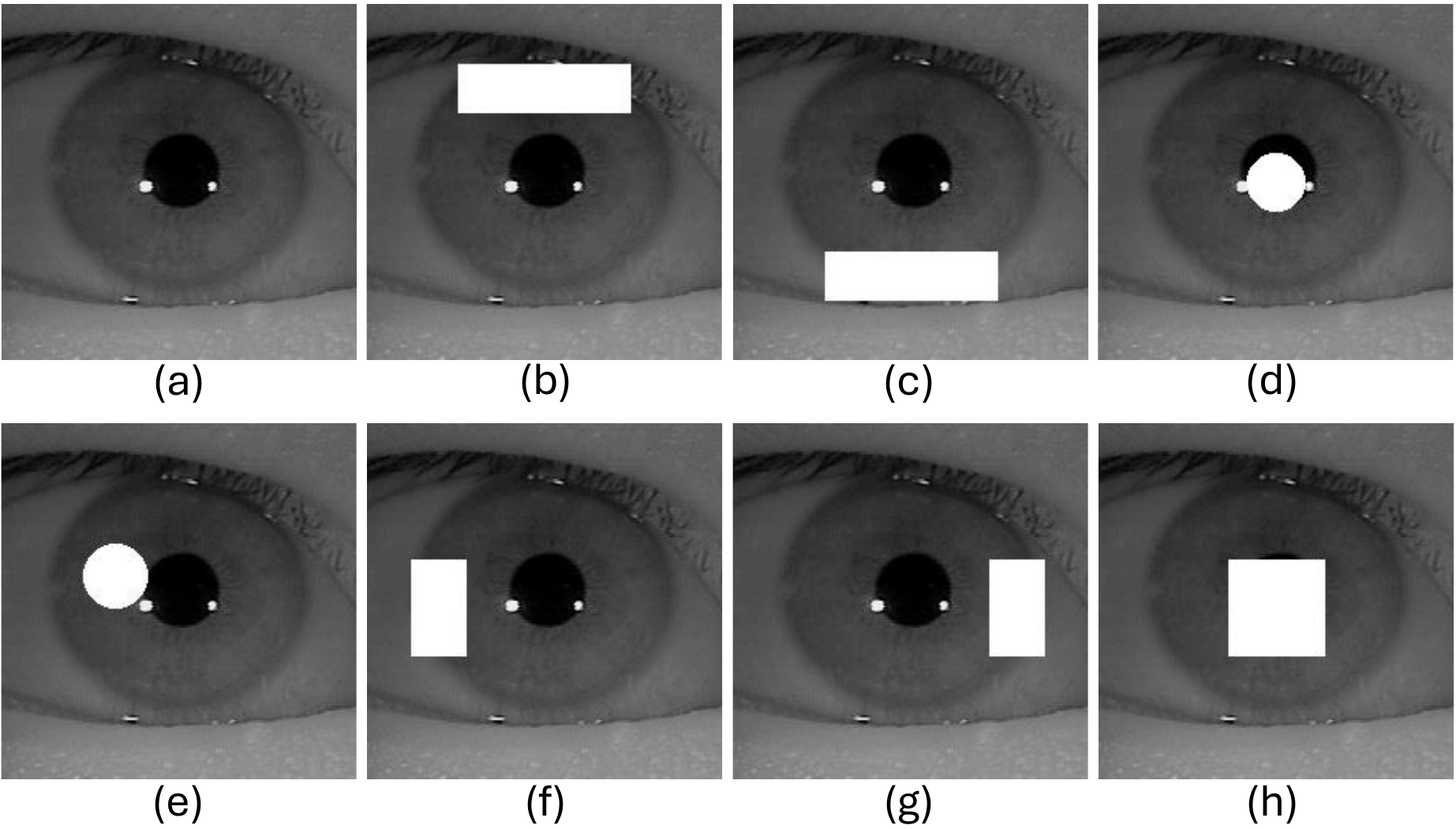}
    \caption{Examples of the synthetic occlusion patterns applied to an iris image. (a) shows the original non-occluded sample, and (b)-(h) show occluded iris images from $c_1$ to $c_7$ types.}
    \label{fig:occlusion_example}
\end{figure}

Since no publicly available iris dataset jointly provides real-world occluded iris images, their corresponding clean ground-truth images, and segmentation masks, we use CASIA-Iris-Thousand as the base dataset and generate rule-based synthetic occlusions for training and evaluation. For each iris image, we retain the original sample $c_0$ and generate seven occluded variants $c_1$--$c_7$, where $c_1$--$c_7$ denote \textit{top\_horizontal\_rectangle}, \textit{bottom\_horizontal\_rectangle}, \textit{central\_small\_circle}, \textit{offcenter\_circle}, \textit{left\_side\_rectangle}, \textit{right\_side\_rectangle}, and \textit{large\_central\_rectangle}, respectively. Overall, the dataset contains 160,000 iris images with occluded versions. Fig. \ref{fig:occlusion_example} illustrates examples of these generated occlusion types.

We additionally evaluate the reconstruction model on the UBIRIS.v2 \cite{proencca2009ubiris} dataset. UBIRIS.v2 contains visible-wavelength iris images acquired at a distance and on the move under less-constrained conditions, including blur, reflections, off-angle views, and partial iris visibility. We apply the same synthetic masks and use each unmasked UBIRIS.v2 image as the reconstruction target. This cross-dataset evaluation assesses model generalization to a different sensor and acquisition domain, rather than reconstruction of naturally occurring occlusions.


\subsection{Data Splits}

\label{sec:protocol}

For occlusion identification and diffusion-based reconstruction, the dataset is divided into 128,000 images for training, 16,000 for validation, and 16,000 for testing. For the recognition task, we adopt a subject-disjoint protocol to evaluate generalization to unseen identities. Specifically, the first 800 subjects and their corresponding synthetic occlusion samples are used for training, while the remaining 200 subjects are reserved for testing only.

\subsection{Training Details}
The occlusion identifier and conditional DDPM use the image-level split in Section~\ref{sec:protocol}. The recognition model is trained on the first 800 subjects, while the remaining 200 are reserved for subject-disjoint evaluation. For all recognition experiments, the first two clean images per subject form the gallery. In the primary protocol, each of the remaining eight images is evaluated under all seven occlusion types, and the best one is selected for evaluation. For the reconstruction-baseline and bootstrap analyses, each subject contributes seven probes, one per occlusion type. Thus, absolute results are compared only within the same protocol. Verification compares probe and gallery embeddings, whereas closed-set identification matches each probe against all enrolled identities. Within each protocol, all reconstruction methods use identical preprocessing, splits, masks, targets, gallery/probe composition, and recognition architecture.

\subsection{Implementation Details}
The occlusion identification module was trained with an initial learning rate of $1\times10^{-4}$ using early stopping. The conditional DDPM module was trained with base channel size 64, embedding dimension 256, dropout 0.1, and exponential moving average decay 0.9999. We used the AdamW optimizer \cite{loshchilov2017decoupled} with a learning rate of $2\times10^{-4}$, weight decay of $1\times10^{-4}$, and batch size 32. The diffusion process was configured with 1000 time steps and a linear noise schedule ranging from $1\times10^{-4}$ to 0.02.

VGG19-HPMNet was trained for up to 300 epochs with batch size 128, learning rate $1\times10^{-4}$, embedding dimension 512, label smoothing 0.1, and weight decay $1\times10^{-4}$. Early stopping was applied with a patience of 20 epochs, and the learning rate was adjusted using a plateau scheduler with a patience of 5 and a decay factor of 0.5. All experiments were conducted using Python 3.9.12, CUDA 12.4, and PyTorch 2.1.0 on two NVIDIA RTX A6000 GPUs, each with 48 GB GDDR6 memory. 

For batch-one inference with \(224 \times 224\) inputs, the occlusion-identification and conditional-reconstruction modules contain 48.28M unique parameters and require approximately 7.26 TFLOPs and \(1.085 \pm 0.017\) s per image using 50-step denoising diffusion implicit model (DDIM) \cite{song2020denoising} sampling. The source code and trained models are available on the Texas State Digital Collections Repository at \url{https://hdl.handle.net/10877/25139}.


\subsection{Evaluation Metrics}

We report multi-class classification metrics, including accuracy, precision, recall, and F1-score for occlusion identification. For reconstruction, we evaluate the visual and perceptual quality of the restored images using the Structural Similarity Index Measure (SSIM), Peak Signal-to-Noise Ratio (PSNR), and Learned Perceptual Image Patch Similarity (LPIPS). For recognition, verification performance is reported using Equal Error Rate (EER) and True Accept Rate (TAR) at fixed False Accept Rate (FAR) operating points, while closed-set identification performance is reported using Rank-1 and Rank-5 identification rates (IR).

\section{Results}
We evaluate the proposed framework in terms of occlusion identification, reconstruction quality, and biometric verification and identification. Reconstruction is compared with TT-GAN \cite{chen2024two}, cGAN \cite{lee2019conditional}, and I3FDM \cite{li2024i3fdm}, while recognition is compared with VGG-16 \cite{simonyan2014very}, ResNet-50 \cite{he2016deep}, Inception-V3 \cite{szegedy2016rethinking}, and ViT \cite{dosovitskiy2020image} under identical training and testing settings.

\subsection{Occlusion Identification Performance}
Tab.~\ref{tab:occlusion_detection} reports the performance of the proposed occlusion identification network for the non-occluded class and seven controlled occlusion classes. The model achieves near-perfect in-distribution performance across all classes, with precision, recall, F1, and accuracy scores close to $100\%$. These results show that the residual CNN can reliably separate the controlled occlusion categories used to condition the downstream diffusion model.

\begin{table}[!t]
\centering
\caption{Performance of the occlusion type identifier module in precision, recall, F1-score, and accuracy.}
\label{tab:occlusion_detection}
\begin{adjustbox}{width=0.47\textwidth}
{\small
\renewcommand{\arraystretch}{1}
\begin{tblr}{
  cells = {c},
  hline{1-2,10} = {-}{},
}
\textbf{Class}    & \textbf{Precision (\%)} & \textbf{Recall (\%)} & \textbf{F1-Score (\%)} & \textbf{Accuracy (\%)} \\
$c_0$ & 99.90          & 99.95       & 99.93         & 99.98        \\
$c_1$ & 100.00         & 100.00      & 100.00        & 100.00        \\
$c_2$ & 100.00         & 100.00      & 100.00        & 100.00        \\
$c_3$ & 100.00         & 99.95       & 99.97         & 99.98        \\
$c_4$ & 99.95          & 100.00      & 99.98         & 99.95        \\
$c_5$ & 100.00         & 100.00      & 100.00        & 100.00        \\
$c_6$ & 100.00         & 99.95       & 99.97         & 99.98        \\
$c_7$ & 100.00         & 100.00      & 100.00        & 100.00        
\end{tblr} 
}
\end{adjustbox}
\end{table}

\subsection{Iris Reconstruction Performance}
Tab.~\ref{tab:reconstruction_quantitative_casia} compares the proposed reconstruction module with TT-GAN, cGAN, and I3FDM under the same CASIA-Iris-Thousand masking protocol. Our proposed method achieves the best performance across all metrics, with a PSNR of $37.68 \pm 3.07$, an SSIM of $0.95 \pm 0.02$, and an LPIPS of $0.05 \pm 0.02$. The closest baseline, cGAN, obtains $36.74 \pm 2.99$, $0.93 \pm 0.02$, and $0.08 \pm 0.04$, respectively. These results indicate that the proposed conditional DDPM provides more accurate and perceptually consistent reconstruction of the masked iris regions.


\begin{table}[h]
\centering
\caption{Comparison of occlusion reconstruction performance with existing approaches using PSNR, SSIM, and LPIPS. Best results are highlighted in bold.}
\label{tab:reconstruction_quantitative_casia}
\begin{adjustbox}{width=0.35\textwidth}
{\small
\renewcommand{\arraystretch}{1}
\begin{tblr}{
  cells = {c},
  row{1} = {font=\bfseries},
  cell{5}{2} = {font=\bfseries},
  cell{5}{3} = {font=\bfseries},
  cell{5}{4} = {font=\bfseries},
  hline{1-2,6} = {-}{},
}
Model  & PSNR~ ↑      & SSIM~ ↑     & LPIPS ↓     \\
TT-GAN \cite{chen2024two} & 23.19 ± 2.86 & 0.90 ± 0.02 & 0.17 ± 0.05 \\
cGAN \cite{lee2019conditional}  & 36.74 ± 2.99 & 0.93 ± 0.02 & 0.08 ± 0.04 \\
I3FDM \cite{li2024i3fdm} & 35.53 ± 2.74 & 0.91 ± 0.03 & 0.06 ± 0.06 \\
Ours   & 37.68 ± 3.07 & 0.95 ± 0.02 & 0.05 ± 0.02 
\end{tblr}
}
\end{adjustbox}
\end{table}

\begin{figure}[ht]
    \centering
\includegraphics[width=\linewidth]{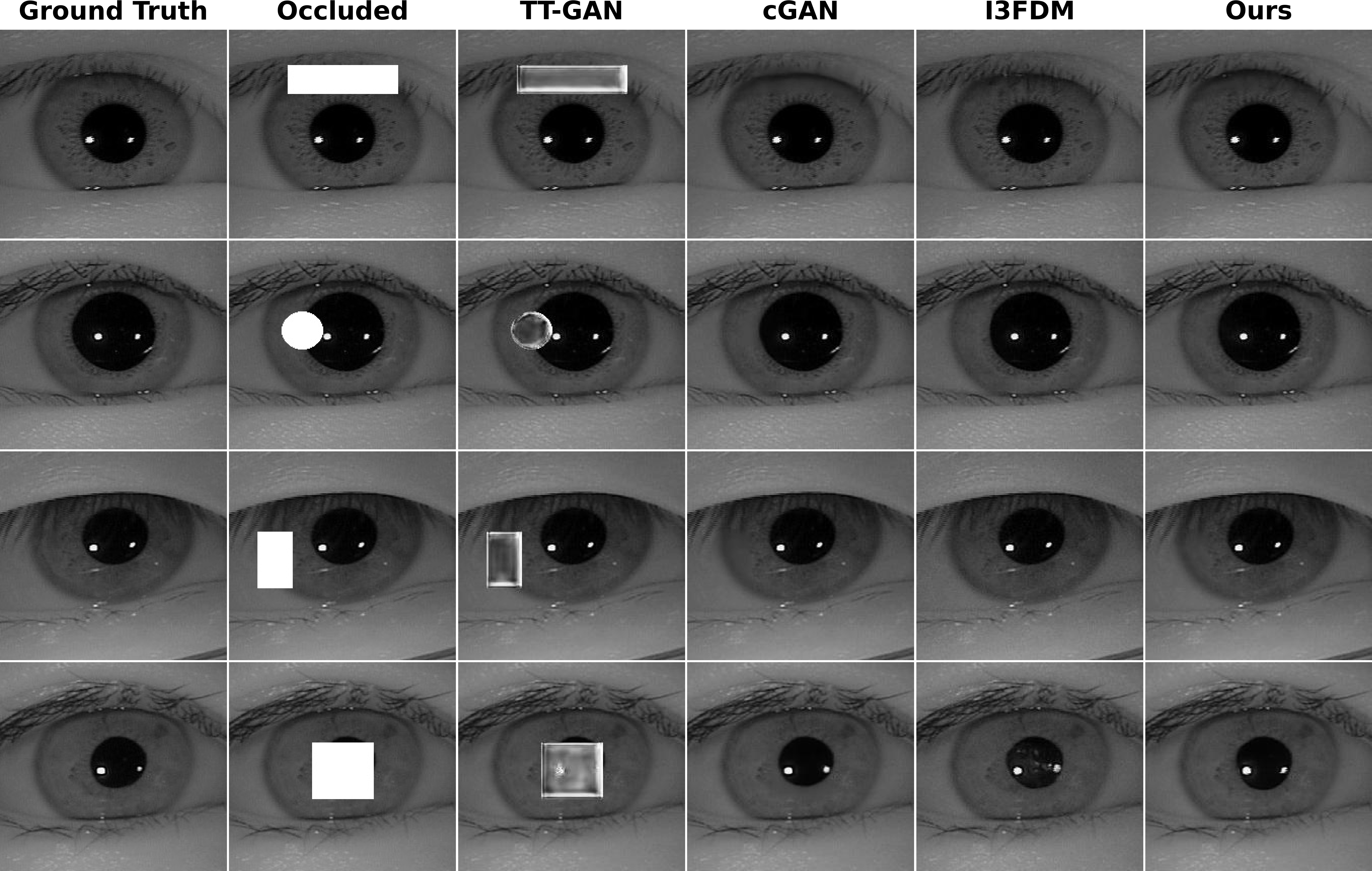}
    \caption{Comparison of reconstructed iris images with existing approaches. Here, ground truth denotes non-occluded images, and the first to fourth rows represent the $c_1$, $c_4$, $c_5$, $c_7$ types of occlusion.}
    \label{fig:occlusion_generated_example}
\end{figure}

Fig.~\ref{fig:occlusion_generated_example} presents qualitative reconstruction examples obtained using identical input masks. TT-GAN often leaves visible artifacts or incomplete restoration near the masked region, while cGAN produces smoother results but may suppress fine iris texture. I3FDM preserves more structure than TT-GAN but exhibits residual inconsistencies for challenging occlusions, particularly large central masks. In comparison, the proposed method produces cleaner transitions and better preserves local texture continuity, consistent with the quantitative results in Tab.~\ref{tab:reconstruction_quantitative_casia}.

To further assess cross-dataset reconstruction, we evaluate the CASIA-Iris-Thousand-trained models on Session~1 of UBIRIS.v2 \cite{proencca2009ubiris} without retraining or fine-tuning. The same synthetic-masking protocol is applied, and each unmasked UBIRIS.v2 image is used as the reconstruction target. As shown in Tab.~\ref{tab:reconstruction_quantitative_ubiris}, the proposed method achieves the best cross-dataset reconstruction performance, with cGAN providing the closest baseline. These results indicate that the proposed conditional reconstruction module retains strong restoration capability even when evaluated on images from a different dataset without additional training.

\begin{table}[h]
\centering
\caption{Cross-dataset reconstruction quality on UBIRIS.v2 without fine-tuning. Best results are highlighted in bold.}
\label{tab:reconstruction_quantitative_ubiris}
\begin{adjustbox}{width=0.35\textwidth}
{\small
\renewcommand{\arraystretch}{1}
\begin{tblr}{
  cells = {c},
  row{1} = {font=\bfseries},
  cell{5}{2} = {font=\bfseries},
  cell{5}{3} = {font=\bfseries},
  cell{5}{4} = {font=\bfseries},
  hline{1-2,6} = {-}{},
}
Model  & PSNR~ ↑      & SSIM~ ↑     & LPIPS ↓     \\
TT-GAN \cite{chen2024two} & 22.40 ± 3.07 & 0.92 ± 0.02 & 0.21 ± 0.07 \\
cGAN \cite{lee2019conditional}  & 38.36 ± 3.42 & 0.95 ± 0.01 & 0.04 ± 0.02 \\
I3FDM \cite{li2024i3fdm} & 15.86 ± 2.74 & 0.94 ± 0.02 & 0.20 ± 0.06 \\
Ours   & 39.00 ± 3.58 & 0.96 ± 0.01 & 0.03 ± 0.01 
\end{tblr}
}
\end{adjustbox}
\end{table}


\begin{table*}[ht]
\centering
\caption{Verification and closed-set identification performance of the compared recognition models with and without reconstruction. Lower EER and higher TAR, Rank-1 IR, and Rank-5 IR indicate better performance. Best and second-best results are shown in bold and italics, respectively.}
\label{tab:maintable}
\begin{adjustbox}{width=0.8\textwidth}
{\small
\renewcommand{\arraystretch}{1}
\begin{tblr}{
  cells = {c},
  row{1} = {font=\bfseries},
  cell{2}{1} = {r=2}{},
  cell{4}{1} = {r=2}{},
  cell{6}{1} = {r=2}{},
  cell{8}{1} = {r=2}{},
  cell{10}{1} = {r=2}{},
  hline{1-2,4,6,8,10,12} = {-}{},
}
Model        & Status             & EER (\%) ↓ & TAR@FAR=1\% (\%) ↑ & TAR@FAR=0.1\% (\%) ↑ & Rank-1 IR (\%) ↑ & Rank-5 IR (\%) ↑ \\
VGG-16       & W/O reconstruction & 3.59     & 94.00            & 87.47              & 90.60               & 94.52               \\
             & W/ reconstruction  & \textit{3.01}     & \textbf{95.77}            & 89.45              & \textit{94.17}               & 96.11               \\
ResNet-50    & W/O reconstruction & 3.49     & 93.44            & 88.90              & 91.83               & 93.52               \\
             & W/ reconstruction  & 3.13     & 94.83            & 87.83              & 89.72               & 95.67               \\
Inception-V3 & W/O reconstruction & 3.77     & 93.87            & 86.39              & 88.75               & 94.81               \\
             & W/ reconstruction  & 3.69     & 93.49            & 85.55              & 88.98               & 94.75               \\
ViT          & W/O reconstruction & 10.82    & 78.04            & 64.33              & 68.18               & 80.52               \\
             & W/ reconstruction  & 5.39     & 89.28            & 81.43              & 84.56               & 92.17               \\
Ours         & W/O reconstruction & 3.29     & 95.39            & \textit{91.75}              & 93.22               & \textit{96.14}               \\
             & W/ reconstruction  & \textbf{2.88}     & \textit{95.72}            & \textbf{92.50}     & \textbf{94.47}    & \textbf{96.58}           
\end{tblr}
}
\end{adjustbox}
\end{table*}

\subsection{Recognition Performance}

\subsubsection{Verification Performance}
Tab.~\ref{tab:maintable} reports verification performance using EER and TAR at FARs of 1\% and 0.1\%. Reconstruction reduces EER across all recognition architectures, with the largest improvement obtained by ViT, whose EER decreases from 10.82\% to 5.39\% and TAR@FAR=0.1\% increases from 64.33\% to 81.43\%. VGG-16 achieves the highest TAR@FAR=1\% of 95.77\%, while the proposed VGG19-HPMNet obtains the lowest EER of 2.88\% and the highest TAR@FAR=0.1\% of 92.50\%. Compared with direct recognition without reconstruction, our proposed reconstruction pipeline reduces EER by 0.41 percentage points and improves TAR by 0.33 and 0.75 percentage points at FARs of 1\% and 0.1\%, respectively, demonstrating its effectiveness under stringent verification conditions.


\subsubsection{Identification Performance}

The closed-set identification results in Tab.~\ref{tab:maintable} show architecture-dependent reconstruction benefits. Rank-1 IR improves for VGG-16, Inception-V3, ViT, and the proposed VGG19-HPMNet, while decreasing for ResNet-50. ViT achieves the largest gains, with Rank-1 and Rank-5 IR increasing from 68.18\% and 80.52\% to 84.56\% and 92.17\%, respectively. The proposed method achieves the best overall performance, reaching 94.47\% Rank-1 IR and 96.58\% Rank-5 IR. Relative to direct recognition, reconstruction improves these metrics by 1.25 and 0.44 percentage points, confirming that the reconstructed images preserve discriminative identity features.


\subsection{Ablation Study}

\subsubsection{Recognition-Stage Baseline Comparison}
We compare all generative approaches under the same recognition architecture, gallery/probe split, and synthetic masking protocol to assess the effect of reconstruction and determine which generative approach better preserves biometric utility. In this comparison, a single sample from all types of occlusion is considered in the probe. As shown in Tab.~\ref{tab:recognition_reconstruction_ablation}, W/O reconstruction obtains 5.16\% EER and 89.32\% Rank-1 IR, while Mean-fill and interpolation improve these results to 4.78\%/90.46\% and 4.73\%/90.34\%, respectively. TT-GAN, an unconditional generative baseline without occlusion-type guidance, performs worst at 5.58\% EER and 89.11\% Rank-1 IR. The proposed conditional diffusion method achieves the lowest EER of 4.72\% and a Rank-1 IR of 90.57\%. Although cGAN obtains a slightly higher Rank-1 IR of 90.71\%, its advantage is only 0.14 percentage points, and its EER is higher at 4.74\%; these differences are further examined using paired bootstrap analysis in Tab.~\ref{tab:paired_bootstrap_ablation}.


\begin{table}[h]
\centering
\caption{Ablation study comparing reconstruction and internal baselines under the same recognition architecture, gallery/probe split, and synthetic masking protocol.}
\label{tab:recognition_reconstruction_ablation}
\begin{adjustbox}{width=0.35\textwidth}
{\small
\renewcommand{\arraystretch}{1}
\begin{tabular}{lcc}
\hline
Method & EER (\%) $\downarrow$ & Rank-1 IR (\%) $\uparrow$ \\
\hline
W/O reconstruction & 5.16 & 89.32 \\
Mean-fill & 4.78 & 90.46 \\
Interpolation & 4.73 & 90.34 \\
TT-GAN & 5.58 & 89.11 \\
cGAN & 4.74 & \textbf{90.71} \\
I3FDM & 4.75 & 90.46 \\
Ours & \textbf{4.72} & 90.57 \\
\hline
\end{tabular}
}
\end{adjustbox}
\end{table}

\subsubsection{Paired Bootstrap Analysis of Recognition Gains}

To evaluate whether recognition-stage differences are stable across held-out subjects, we perform a paired subject-level bootstrap analysis between the proposed method and each baseline. In each bootstrap trial, the same held-out test subjects are sampled with replacement for both methods, all occlusion types for the sampled subjects are retained, and the differences in EER and Rank-1 IR are recomputed. The reported intervals are $95\%$ bootstrap confidence intervals over the paired method differences. As shown in Tab.~\ref{tab:paired_bootstrap_ablation}, the proposed method provides stable improvements over direct recognition without reconstruction, reducing EER by $0.50$ percentage points with a $95\%$ confidence interval of $[-0.89,-0.19]$ and improving Rank-1 IR by $1.27$ percentage points with a confidence interval of $[0.61,2.04]$. A stable improvement is also observed over TT-GAN. For Mean-fill, Interpolation, cGAN, and I3FDM, the confidence intervals include zero, indicating comparable recognition performance rather than statistically conclusive superiority. Therefore, the proposed method should be interpreted as clearly better than direct recognition and TT-GAN under this evaluation while remaining competitive with the stronger filling and generative reconstruction baselines.

\begin{table}[h]
\centering
\caption{Paired subject-level bootstrap comparisons under the pooled seven-probe protocol. Positive $\Delta$Rank-1 and negative $\Delta$EER indicate improvement over the baseline; values are mean differences [95\% CI] in percentage points.}
\label{tab:paired_bootstrap_ablation}
\begin{adjustbox}{width=0.42\textwidth}
{\small
\renewcommand{\arraystretch}{1}
\begin{tabular}{lcc}
\hline
Comparison & $\Delta$EER (\%) $\downarrow$ & $\Delta$Rank-1 IR (\%) $\uparrow$ \\
\hline
Ours -- W/O reconstruction & -0.50 [-0.89, -0.19] & +1.27 [0.61, 2.04] \\
Ours -- Mean-fill & -0.14 [-0.42, 0.09] & +0.12 [-0.43, 0.61] \\
Ours -- Interpolation & -0.01 [-0.20, 0.22] & +0.26 [-0.14, 0.71] \\
Ours -- TT-GAN & -0.78 [-1.22, -0.31] & +1.48 [0.71, 2.32] \\
Ours -- cGAN & -0.01 [-0.15, 0.15] & -0.15 [-0.39, 0.04] \\
Ours -- I3FDM & -0.08 [-0.49, 0.36] & +0.12 [-0.54, 0.71] \\
\hline
\end{tabular}
}
\end{adjustbox}
\end{table}

\subsubsection{Effect of Occlusion-Type Conditioning in DDPM}
Tab.~\ref{tab:ablation1} presents the effects of reconstruction and class conditioning. An unconditioned DDPM improves EER/Rank-1 IR from $3.29\%/93.22\%$ to $3.00\%/93.81\%$, while occlusion-type conditioning further improves them to $2.88\%/94.47\%$. Thus, class guidance provides an additional 0.12-percentage-point EER reduction and 0.66-percentage-point Rank-1 gain.

\begin{table}[h]
\centering
\caption{Effect of reconstruction and occlusion-type conditioning.}
\label{tab:ablation1}
\begin{adjustbox}{width=0.42\textwidth}
{\small
\renewcommand{\arraystretch}{1}
\begin{tblr}{
  cells = {c},
  row{1} = {font=\bfseries},
  hline{1-2,5} = {-}{},
}
Reconstruction & Class condition & EER (\%) $\downarrow$ & Rank-1 IR (\%) $\uparrow$\\
- & - & 3.29 & 93.22 \\
\checkmark & - & 3.00 & 93.81 \\
\checkmark & \checkmark & \textbf{2.88} & \textbf{94.47}
\end{tblr}
}
\end{adjustbox}
\end{table}

\subsubsection{Effect of HPM in Recognition}
Tab.~\ref{tab:ablation2} shows that reconstruction alone improves the plain VGG-19 baseline from $3.36\%/88.33\%$ to $3.05\%/93.80\%$ EER/Rank-1 IR, while HPM alone reaches $3.29\%/93.22\%$. Their combination performs best at $2.88\%/94.47\%$. These results indicate that reconstruction provides the larger contribution to individual recognition, while HPM adds a complementary gain when combined with part-wise feature aggregation.



\begin{table}[h]
\centering
\caption{Effect of reconstruction and HPM on recognition.}
\label{tab:ablation2}
\begin{adjustbox}{width=0.38\textwidth}
{\small
\renewcommand{\arraystretch}{1}
\begin{tblr}{
  cells = {c},
  row{1} = {font=\bfseries},
  hline{1-2,6} = {-}{},
}
Reconstruction & HPM & EER (\%) ↓ & Rank-1 IR (\%) ↑\\
-             & -  & 3.36     & 88.33          \\
\checkmark            & -  & 3.05     & 93.80          \\
-             & \checkmark & 3.29     & 93.22          \\
\checkmark            & \checkmark & \textbf{2.88}     & \textbf{94.47}          
\end{tblr}
}
\end{adjustbox}
\end{table}

\section{Discussion}
The results support three main observations. First, explicit reconstruction of occluded iris regions can improve downstream biometric recognition, but the improvement depends on the recognition architecture and operating point. In the main comparison (Tab. \ref{tab:maintable}), reconstruction reduces EER across all evaluated backbones, while gains in TAR and identification vary by model. For VGG19-HPMNet, reconstruction reduces EER from $3.29\%$ to $2.88\%$, improves TAR@FAR=$1\%$ from $95.39\%$ to $95.72\%$, improves TAR@FAR=$0.1\%$ from $91.75\%$ to $92.50\%$, and increases Rank-1 IR from $93.22\%$ to $94.47\%$.

Second, the ablation studies (Tab. \ref{tab:ablation1} and Tab. \ref{tab:ablation2}) show that the recognition gain does not come from reconstruction alone. The unconditioned DDPM does not improve performance much over direct recognition, whereas the occlusion-type-conditioned DDPM achieves the best result. This suggests that explicit occlusion-type information helps guide the reconstruction process. The HPM ablation further shows that part-wise horizontal feature aggregation provides a strong recognition benefit and that reconstruction and HPM are complementary.

Third, the reconstruction-baseline (Tab. \ref{tab:recognition_reconstruction_ablation}) and bootstrap analyses (Tab. \ref{tab:paired_bootstrap_ablation}) provide a conservative interpretation of biometric utility. The key takeaway is that the proposed method achieves the lowest EER among the evaluated reconstruction choices and provides bootstrap-supported improvements over direct recognition and TT-GAN. However, its differences relative to Mean-fill, Interpolation, cGAN, and I3FDM are not statistically conclusive because the corresponding confidence intervals include zero. Therefore, the proposed method should be described as improving recognition over direct matching and weaker reconstruction baselines, while remaining competitive with stronger reconstruction alternatives. Its main advantage is the combination of strong reconstruction quality, explicit masked-region restoration, and occlusion-type-conditioned diffusion.

\section{Limitations}

The evaluation relies on controlled synthetic occlusions and predefined categories, so it may not fully represent natural iris degradations. Cross-dataset results provide preliminary evidence of generalization, and paired clean and occluded images enable controlled mask localization. Because biometric utility is evaluated on fully reconstructed images, generated regions should be regarded as plausible estimates rather than exact recovery of missing iris texture. Future work will investigate reconstructed-region identity cues, challenging and natural occlusions, and faster alternatives to DDPM sampling.

\section{Conclusions}
We presented an occlusion-aware iris recognition framework combining controlled occlusion-type identification, conditional diffusion-based reconstruction, and the VGG19-HPMNet module. The proposed reconstructor achieved the best image-quality metrics among the evaluated baselines and improved verification and closed-set identification under synthetic occlusion. Ablations showed complementary benefits from class conditioning and HPM, while bootstrap analysis supported comparable gains over reconstruction baselines. Overall, the results support occlusion-guided reconstruction as an effective strategy for improving iris recognition under partial occlusion. Future work will focus on automatic mask prediction, broader occlusion handling, naturally occluded evaluation, and faster diffusion inference.

{\small
\bibliographystyle{ieee}
\bibliography{egbib}
}

\end{document}